# Utility Elicitation as a Classification Problem

**Urszula Chajewska** and **Lise Getoor**
Computer Science Department
Stanford University
Stanford, CA 94305-9010
*urszula@cs.stanford.edu*
*getoor@cs.stanford.edu*

**Joseph Norman** and **Yuval Shahar**
Stanford Medical Informatics
Stanford University School of Medicine
Stanford, CA 94305-5479
*norman@smi.stanford.edu*
*shahar@smi.stanford.edu*

## Abstract

We investigate the application of classification techniques to utility elicitation. In a decision problem, two sets of parameters must generally be elicited: the probabilities and the utilities. While the prior and conditional probabilities in the model do not change from user to user, the utility models do. Thus it is necessary to elicit a utility model separately for each new user. Elicitation is long and tedious, particularly if the outcome space is large and not decomposable. There are two common approaches to utility function elicitation. The first is to base the determination of the user's utility function solely on elicitation of qualitative preferences. The second makes assumptions about the form and decomposability of the utility function. Here we take a different approach: we attempt to identify the new user's utility function based on classification relative to a database of previously collected utility functions. We do this by identifying clusters of utility functions that minimize an appropriate distance measure. Having identified the clusters, we develop a classification scheme that requires many fewer and simpler assessments than full utility elicitation and is more robust than utility elicitation based solely on preferences. We have tested our algorithm on a small database of utility functions in a prenatal diagnosis domain and the results are quite promising.

## 1 INTRODUCTION

Probabilistic systems, such as Bayesian networks (Pearl 1988) and influence diagrams (Howard and Matheson 1984) are a major research focus and have been gaining popularity in a number of application domains. Many systems based on them are now in use. Some of these systems, especially those used in medical domains, are designed to give advice to large numbers of users. Often these users do not agree on their preferences in a given decision context. Therefore, in order to recommend appropriate actions, we need to elicit a utility function not once, as a part of creating a model, but many times—once for each user. This can be an extremely long and tedious process, particularly if the outcome space is large and the utility function is not easily decomposable into independent components.

Utility elicitation has been studied extensively in the area of decision analysis (DA) (Luce and Raiffa 1957; Keeney and Raiffa 1976; Howard 1984a; Howard 1984b). Recently it has started to receive attention in medical informatics (Heckerman et al. 1992; Farr and Shachter 1992; Hornberger et al. 1995) and artificial intelligence (AI) (Ha and Haddawy 1997; Linden et al. 1997; Boutilier et al. 1997). Most of the research concentrates on decomposing utility functions, taking advantage of various assumptions of independence between the attributes. Decomposed functions are easier to elicit and can allow more efficient reasoning procedures. Common approaches restrict attention to additive models (Hornberger et al. 1995; Linden et al. 1997), and partial elicitation of models (Ha and Haddawy 1997).

In many cases, however, attributes are preferentially dependent and thus assumptions of decomposability are suspect. An incorrect assumption of decomposability can adversely affect our choice of actions. This problem suggests that we should perform full utility elicitation. However, outcome spaces can be very large and eliciting full utility functions from every user may be infeasible.

Yet, there is still hope for effective elicitation of users' utility functions. Quite often, there are only a few qualitatively different utility functions and we can partition most users' utility functions into classes with very similar functions within each group. Having these clusters of utility functions defined and knowing their prevalence in the population can guide the process of utility elicitation from a new user.

In our approach, we begin with a database of fully-specified utility functions[1]. From this database, we cluster the utility functions in such a way that in a given context, there is some strategy that is close to optimal according to all the functions in that cluster. This strategy is the optimal strategy for some function in the cluster and we identify that function with the cluster. Next, we build a decision tree for classifying the utility functions according to their clusters.

---

[1] The appreciation of the importance of individual preferences is growing in the medical community. Some medical informatics centers are collecting users' preferences for medical decision analysis. We have reasons to hope that such studies will become common and thus databases of patient utility functions will soon be available for many domains.

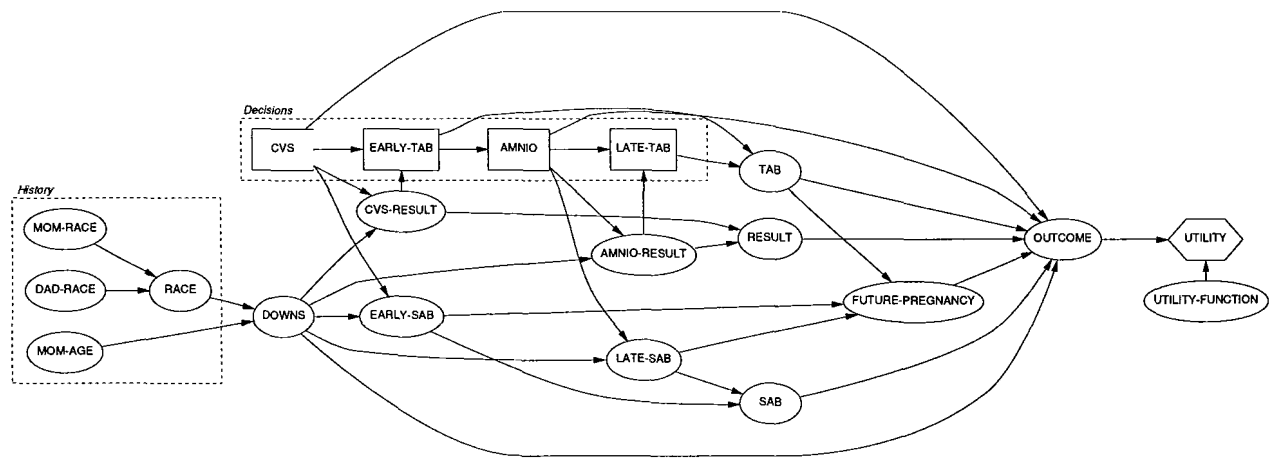

Figure 1: A simplified version of PANDA model.

When we are presented with a new user, we use this decision tree to find a suitable cluster and the utility function that is associated with it. The types of questions posed to the user during this elicitation will be easier for them to answer than the questions required during full elicitation. In addition, the number of questions required to classify a new utility function should be significantly fewer than the full utility function elicitation would require.

The domain we focus on is prenatal testing. We are using a simplified version of the model developed by the PANDA project at Stanford Medical Informatics[2]. PANDA is a loose acronym for "prenatal testing decision analysis." PANDA uses knowledge gained from many studies and from practicing clinicians to advise patients on which prenatal diagnostic tests they should choose during their pregnancies. The full model includes data about six major diseases which can be diagnosed before birth, along with their prevalence and severities. It considers four tests used to diagnose these diseases. These tests have different sensitivities, specificities, costs, and health risks. In this paper, we use a simpler model, shown in Figure 1. Our simplified version considers only one disease—Down's syndrome—and two tests which can diagnose it—chorionic villus sampling (CVS) and amniocentesis (AMNIO).

In the real world, the decision about the choice of tests is rarely easy. The patient's risk for having a child with a serious disease depends on the mother's age, child's sex and race, and the family history. Some tests are not very accurate; others carry a significant risk of inducing miscarriages. Both miscarriage (SAB) and elective termination of the pregnancy (TAB) can affect the woman's chances of conceiving again.

The outcomes in our models have many attributes: the inconvenience and expense of fairly invasive testing, the possibility of test-induced miscarriage, knowledge about the health of the child early in the pregnancy, the possibility of future conception, and the actual health of the child. A recent study (Kuppermann et al. 1997) showed that these attributes are highly correlated and the utility of an outcome cannot be predicted from the utilities of the individual attributes. For example, consider the attributes "future pregnancy" and "miscarriage". While we can generally assume that a woman would like to conceive again following a miscarriage, and thus the attribute "future pregnancy" will be preferred to its negation, we can make no such assumption when a miscarriage has not occurred. Our initial analysis of the model revealed the considerable influence of the utility function (especially the patient's attitude towards the risk of having a child with a serious disease and toward a miscarriage) on the optimal choice of actions.

In the following section, we review some concepts from utility theory in the context of this example. In Section 3, we describe our approach to identifying clusters of utility functions. In Section 4, we show how these clusters are used to classify a new user's utility function. In Section 5, we present the results of our experiments on a small database of utility functions. In Section 6, we review some of the related work. Finally, in Section 7, we discuss potential benefits of our approach and future directions for this work.

## 2   REVIEW: UTILITY THEORY

The principle of maximizing expected utility has long been established as the guide to making rational decisions. The axioms of utility theory, which are stated in terms of constraints on preferences, imply the existence of utility functions (von Neumann and Morgenstern 1947; Savage 1954; Luce and Raiffa 1957). Let $O$ be a set of possible outcomes $\{o_1,\ldots,o_n\}$. The outcomes are also called situations (in AI literature) or consequences (in DA literature). For our model, the possible outcomes include sequences of events;

---

[2]See http://smi-web.stanford.edu/projects/panda/.

there are 22 outcomes in the simplified version. The set of possible strategies, $S = \{s_1, \ldots, s_m\}$, contains all possible decision sequences (conditional plans), such as: "take CVS; if the result is negative, do not take any more tests; otherwise, take amniocentesis; if the result is negative, continue; if not, terminate the pregnancy." The patient's history, $h_k$, is an instantiation of the observable variables in the model representing information specific to the patient and her pregnancy: patient's age, child's sex and race, family history of diabetes, etc. (Since we are only considering one disease in our simplified model, namely Down's syndrome, we are able to reduce the number of history variables to only one relevant to this disease: mother's age.)

The given decision strategy together with the patient's history induces a probability distribution over the set of outcomes $P(\mathbf{O}|\mathbf{H},\mathbf{S})$. Given a probability distribution $P$ and the user's utility function defined over the outcomes, $U(\mathbf{O})$, we can compute the expected utility for the given patient and the chosen strategy:

$$\text{EU}(s|h) = \sum_o P(o|h,s)U(o).$$

Von Neumann and Morgenstern (1947) developed an approach to utility elicitation based on measuring the strength of a person's preference for an outcome by the risks he or she is willing to take to obtain it. Consider three outcomes $o_1$, $o_2$, and $o_3$ and a user with the preference ordering $o_1 \succ o_2 \succ o_3$. If he or she is offered a choice between $o_2$ for sure and a gamble in which $o_1$ will be received with probability $\pi$ and $o_3$ with probability $(1 - \pi)$, then, according to the theory, there exists a value of $\pi$ for which the user will be indifferent. The outcome $o_2$ can then be assigned the utility value $\pi U(o_1) + (1 - \pi)U(o_3)$.

The utility function was shown by von Neumann and Morgenstern (1947) to be uniquely determined up to an increasing linear transformation, i.e., for any utility function $U(O)$ and constants $a$ and $b$, such that $a > 0$, $aU(O) + b$ is also a utility function encoding the same preferences. The constant $a$ changes the scale of the utility function. The constant $b$ changes its zero point.

In order to compare two utility functions, we have to make sure that they are normalized, i.e., their zero points and scales are the same. Usually this is done by finding two endpoints of the scale—the best and worst possible outcomes, $o_\top$ and $o_\bot$—and assigning them the values of 1 and 0 respectively. However, the worst outcome in any given set does not necessarily have the same value for every person. Thus it is common practice to include the death of the decision maker in the set of outcomes, on the presumption that its value is equally abysmal for everyone.

In this paper, we will assume that the utility functions in our database have been normalized. In the PANDA domain, the two outcomes chosen for the endpoints are $o_\top$, the birth of a healthy baby following a healthy pregnancy with full knowledge throughout that the baby is not affected by any disease, and $o_\bot$, the death of the pregnant woman herself; the utility functions take the values in the interval $[0, 1]$.

## 3 IDENTIFYING CLUSTERS OF UTILITY MODELS

We assume that we have a database of $N$ normalized utility functions over our outcome space $O$, where $|O| = D$. Our data points—the utility functions—are represented as vectors of values, one value for each outcome, $\{u(o_1), \ldots, u(o_D)\}$. We also assume that the utility functions of our new users will be drawn from the same distribution as those in our database.

In order to create the utility clusters, we can use any of the popular clustering algorithms. The goal is to divide a set of data points into non-overlapping groups, or clusters, of points, where points in a cluster are "more similar" to one another than to points in other clusters. When a dataset is clustered, every point is assigned to some cluster, and every cluster can be characterized by a single reference point which we will call its *prototype*.

### 3.1 CHARACTERIZING SIMILARITY BETWEEN UTILITY FUNCTIONS

A key component of any clustering algorithm is a notion of distance between points. The simplest approach would be to treat all utility functions as vectors of values, one for each outcome, and use the Euclidean distance between them as the distance measure. In essence this approach gives each outcome equal weight. It would be a mistake, however, to do this. Not all outcomes are equally probable, thus the values attached to different outcomes by a utility function contribute differently to the expected utility value. For example, the probability of having a miscarriage is much smaller than the probability of having a healthy child and thus the changes in utilities for these outcomes do not affect the value of the expected utility equally.

How can we resolve this problem? Since we are clustering the utility functions in the context of a specific decision problem, we can take advantage of the information contained in our model. The quantity we want to minimize is the difference between the expected utility of a strategy we would choose for our user if we elicited her full utility function and the expected utility of the strategy we would choose for her based on our algorithm.

To specify it more formally, we begin by defining the expected utility of a particular strategy $s$ with respect to a particular utility function $u_i$ and a particular history $h_k$:

$$\text{EU}_{u_i}(s|h_k) = \sum_{o_l} P(o_l|s,h_k)u_i(o_l)$$

where $o_l$ ranges over possible outcomes. We can then look at the *best strategy* for a particular utility function and history,

$$s^*_{u_i|h_k} = \text{argmax}_s \text{EU}_{u_i}(s|h_k).$$

In the following, let $u_p$ be the utility function for some prototype $p$ and let $s^*_{u_p|h_k}$ be the best strategy for this prototype.

Eventually, we will be giving advice to new users based some cluster's prototype. We would not like the result to differ significantly from what users could expect based on the full utility elicitation. In order to compute this difference, we need to consider two possibly different strategies: the strategy that we will pick for $u_p$, $s^*_{u_p|h_k}$, and the strategy that we would pick for the true utility function, $\tilde{u}$, $s^*_{\tilde{u}|h_k}$. We will evaluate both of these strategies for a particular history $h_k$ using $\tilde{u}$.

**Definition 3.1:** The *Utility Loss* (UL) for a utility function $\tilde{u}$ with respect to a utility function $u_p$ and a history $h_k$ in the context of a given decision model $M$ is

$$\mathrm{UL}(\tilde{u}, u_p|h_k) = \mathrm{EU}_{\tilde{u}}(s^*_{\tilde{u}|h_k}) - \mathrm{EU}_{\tilde{u}}(s^*_{u_p|h_k})$$

where the expected utility is defined with respect to $M$. ∎

We will use this as the score we wish to minimize. Note that this measure is not symmetric: the UL of a utility function $u_i$ with respect to another utility function $u_j$ may be very different than the UL of $u_j$ with respect to $u_i$. This asymmetry matches the intended use of the measure. We will use one prototype function to advise many users.

Based on our measure of the UL, we define the distance between two utility functions as follows:

**Definition 3.2:** The *distance* between two utility functions $u_i$ and $u_j$ with respect to a history $h_k$ is defined as

$$d(u_i, u_j|h_k) = \frac{\mathrm{UL}(u_i, u_j|h_k) + \mathrm{UL}(u_j, u_i|h_k)}{2}$$ ∎

Note that our distance measure is not a metric; while it is symmetric, it does not satisfy the triangle inequality.

We could create clusters based on averaging over histories, by minimizing the following distance:

$$d(u_i, u_o) = \sum_{h_k} P(h_k) \cdot d(u_i, u_o|h_k)$$

This approach may be satisfactory when using the strategy that is very good in most cases and does not have serious consequences even when we are faced with an unlikely scenario. However, in a medical setting, where it is imperative to follow the correct strategy in the case of unlikely situation, this approach is not appropriate.

Instead, in our algorithm, we are creating clusters for a particular history. We can do this in two ways: online and offline. In the online version, we start the process when we are presented with a new user whose utility function we want to classify. At that point, we know the particular history $h_k$ that we are interested in, and we create the clusters relative to that history. Alternatively, we may do this offline, and create clusters for each potential history. The choice will depend on the size of the problem, the response requirements for the online utility elicitation and the storage availability. Note that in either case, we are able to use the entire database to build our clusters because we assume

```
Procedure Clustering

Inputs:   N number of data points
          k maximum number of classes
          u_i utility functions to be clustered; i = 1..N
          h_k current history
Outputs:  C_j clusters; j = 1..k and their prototypes u_{C_j}

For each u_i
    Put u_i in a separate cluster C_i
    Add C_i to the list of clusters L
For each C_i and for each C_j
    d(C_i, C_j) = (UL(u_i, u_j|h_k) + UL(u_j, u_i|h_k))/2
Repeat
    Find clusters C_r, C_s that are most similar
    Merge C_r with C_s to form a new cluster C_t
    Remove C_r and C_s from L
    For each C_i on L
        d(C_t, C_i) = (|C_r|d(C_r,C_i)+|C_s|d(C_s,C_i))/(|C_r|+|C_s|)
    Add C_t to L
until the number of clusters is less than or equal to k
For each cluster C_n
    Pick a representative u_{C_n} with the lowest score
        Score(u_i) = Σ_{u_j∈C_n} UL(u_j, u_i|h_k)
```

Figure 2: Hierarchical Agglomerative Clustering Algorithm.

that the utility function is independent of the patient history. This lack of data fragmentation is one of the benefits of this approach.

### 3.2 CLUSTERING THE UTILITY FUNCTIONS

Clustering algorithms come in two general flavors, partitioning methods (such as $k$-means) and hierarchical methods (Willett 1988). In general, hierarchical methods are faster than partitioning methods. We are concerned about the efficiency of our algorithm, since it may be done in an online setting. Therefore, we chose a hierarchical agglomerative clustering algorithm. Another benefit of a hierarchical method is that it allows us to tradeoff explicitly between the number of clusters and the similarity score.

The algorithm starts by putting every data point in a separate cluster. Then, it computes the distances between every two clusters. It finds the two closest clusters and combines them into one. It continues to merge clusters until some stopping criterion is met.

There are several ways to define the distances between clusters containing more than one element. We use one of the standard definitions, the *group average link method*, which uses the average values of the pairwise links within the potential new cluster to determine similarity.

After the clustering, we choose a prototype (cluster representative) for every cluster by picking the utility function with the lowest score

$$\mathrm{Score}(u_i) = \sum_{u_j \in \mathrm{cluster}} \mathrm{UL}(u_i, u_j|h_k)$$

## 4   CLASSIFYING UTILITY MODELS

Given that we have found our $k$ clusters, we would like to find the cluster to which a new user's utility is most likely to belong, and use the prototype utility function for that cluster to determine the recommended strategy for our new user. Once we have clustered the data, we can label each utility function in the database with the cluster to which it was assigned in the clustering phase. Thus the task of identifying a cluster prototype for the new user is a classification problem. At this point, we could apply a nearest neighbor or case-based approach to find a cluster label for the new user. The problem is that we would still need to elicit the user's full utility function in order to do the nearest neighbor calculation, which would defeat the purpose of our endeavor. Instead, we consider building a decision tree for the labeled database. We construct the tree by choosing tests and recursively splitting the database into partitions based on the outcomes of the tests. We keep splitting the partitions until the labels of the utility functions in each partition are largely from the same cluster. These will be the leaves of the decision tree. In order to classify a new user, we begin at the root of the decision tree, and ask the user to answer the test at each node in the decision tree. We traverse a path down the tree until we reach a leaf node. We then classify the new user's utility function according to the labels of the database utility functions that are at that leaf. Decision trees are particularly appropriate in this context because of the ease of human interpretation. More importantly, we will see that the questions required of users are of a form that are easier to answer than standard gamble questions.

### 4.1   BUILDING THE DECISION TREE

The key components of decision tree induction algorithms are: the types of splits considered, the splitting criteria, or how the best split is chosen, and stopping or pruning rules. We describe each of these in turn.

There are $N$ utility functions in our database and $D$ features in each utility vector. Each of the features is a real value between 0.0 and 1.0 that is the normalized utility for a particular outcome in the outcome space. There are two types of splits we allow in the tree: preference splits and feature splits. Feature splits are of the form "Is outcome $o_i$ preferred to the standard lottery, $[c, o_\top; (1-c), o_\bot]$?". In decision tree algorithms that allow real valued features, this is the standard type of split considered. Preference splits are of the form "Is outcome $o_i$ preferred to outcome $o_j$?" Preference splits are a special type of linear combination splits. These linear combination splits are supported in some of the more sophisticated decision tree packages, such as CART (Breiman et al. 1984); here, by using the domain knowledge and looking only at preference splits, we need to consider only a small subset of all the possible linear combination splits.

If we were doing full model elicitation, users would need to answer questions of the form: "For what value of $c$ is the user indifferent between outcome $o_i$ and the standard lottery, $[c, o_\top; (1-c), o_\bot]$?". The questions required for the feature splits and preference splits are each easier to answer. Rather than having to specify a value for $c$, the user just has a yes or no question to answer.

How do we compute the best split? There are many splitting rules considered in the literature. Each in some sense measures the impurity of a child node in the tree relative to its parent. The purity of a node $n$ is a measure of the concentration of labels at that node in the tree. A node is pure if the labels of all of the examples at that node are the same. The impurity reaches its maximum if the distribution of labels is uniform. The most common measure of impurity is *entropy*:

$$I(n) = -\sum_{i=1}^{k} \hat{p}_i \log \hat{p}_i,$$

where $\hat{p}_i$ is an estimate of the probability of having cluster label $i$ given that we are at node $n$ in the decision tree. $\hat{p}_i$ is simply the number of examples at node $n$ having cluster label $i$ divided by the total number of examples at node $n$.

Once we have chosen our measure, we can compute the gain in purity for a particular split $s$. It is the impurity of the node $n$ minus the impurity of each of its children, $n_l$ and $n_r$, weighted by the estimated probability of being at each of the children given the split $s$:

$$Gain = I(n) - P(n_l|s)I(n_l) - P(n_r|s)I(n_r).$$

Our tree building algorithm is a greedy algorithm that chooses the split that has the greatest *Gain* at each step. Thus we compute, for each preference split and each feature split, the gain with that split. There are $D(D-1)/2$ preference splits to consider, each of the form $o_i \succ o_j$, for each $i$, $1 \leq i < D$, and each $j > i$. For each feature, there are at most $N$ feature splits to consider, each of the form $u(o_i) \leq c$, where $i$ is the feature we are splitting on and $c$ is the split point. We need only consider as split points observed values of that feature for some utility function in our database.

We disallow a feature split if the values immediately to the left and right of the split are very close. We prefer to find "gaps" in the values for the feature. We want the questions to be easy for the user to answer. If the split value is too close to the user's utility value, the question may be very difficult for her to answer and her answer would not be reliable. For example, if we split on the utility of the outcome "No tests, child with Down's syndrome" at the value 0.975, and the user's utility is quite close to that value, she will have a hard time answering the question. The appropriate threshold can be determined experimentally. Note that such a threshold will introduce a slight bias towards preference questions. This is quite fortunate, since preference questions are much easier for the user to answer.

There are also many stopping criteria to consider. Here we stop when all the utility functions have the same cluster label.

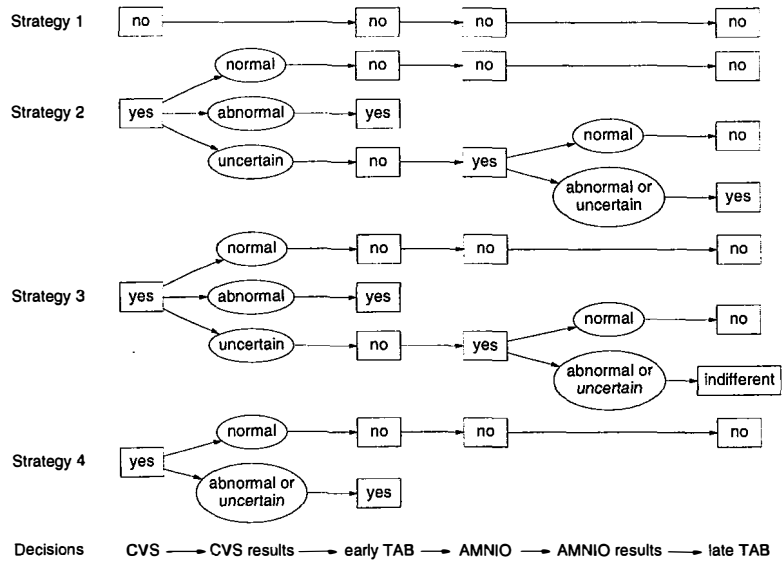

Figure 3: Optimal Strategies

### 4.2 USING THE TREE TO CLASSIFY NEW USERS

Note that in general, for full utility elicitation, there are $|O|$ lottery questions that must be asked of the form "For what value of $c$ is the user indifferent between outcome $o_i$ and the standard lottery, $[c, o_\top; (1-c), o_\bot]$?". In our decision tree classification, there will be at most $d$ questions asked, where $d$ is the maximum depth of the tree, and each of the questions will be a preference question of the form, "Is outcome $o_i$ preferred to outcome $o_j$?" or, for a feature split a question of the form, "Is outcome $o_i$ preferred to the standard lottery, $[c, o_\top; (1-c), o_\bot]$?", for fixed $c$.

How is this procedure helping us to advise new users? First, we collect the history data $h_k$ from the user. We then find the appropriate clustering for that history, and classify the new user's utility function. When a user's utility function is determined to belong to a given cluster characterized by the prototype $p_i$, we find the best strategy for the patient's history $h_k$ and the prototype's utility function $u_{p_i}$. Thus we find a nearly-optimal strategy for the user with only a small number of question.

Decision trees have many well known advantages. In our context, an important advantage is the ease of handling both preference splits and feature splits in a single framework. This allows us to seamlessly integrate the two types of questions in a well-founded manner based on information theoretic principles. Another important advantage is the human interpretability of a decision tree; practitioners can look at the tree and see if the classifications it provides make sense. Finally, as mentioned earlier, the key advantage that we are exploiting for elicitation is that the questions required of the users are significantly easier for them to answer.

Decision trees also have many disadvantages. All techniques for building decision trees rely on greedy strategies, because finding the optimal decision tree is intractable. This leads to instability and high variance in the algorithms. A small perturbation in the input data can lead to quite different decision trees. Our approach of first clustering the data, before building the decision tree for the data, should help us overcome this problem. We have not explored this issue in depth, but see it an interesting area for further work.

There are many related data structures such as regression trees and KD-trees. It is possible to reformulate our approach of clustering based on UL and building a classification tree from the clusters, into a modification of the appropriate regression tree or KD-tree algorithms, where we use UL to compute the distances and we consider splits that are equivalent to our preference and feature splits. The resulting algorithms are equivalent in terms of time and space complexity.

## 5 EMPIRICAL RESULTS

We ran our algorithm with the simplified PANDA model and a database of utility functions that were collected by Miriam Kuppermann of UCSF/Mount Zion Medical Center (1997). There were 70 utility functions in the database. Of these, many had missing values. In the experiments reported, we ran on the 55 utility functions with no missing values.

We considered four different patient histories, corresponding to the age of the woman: TEEN, 25YO, 35YO, and 45YO. We also considered an average history, AVE, as a baseline.

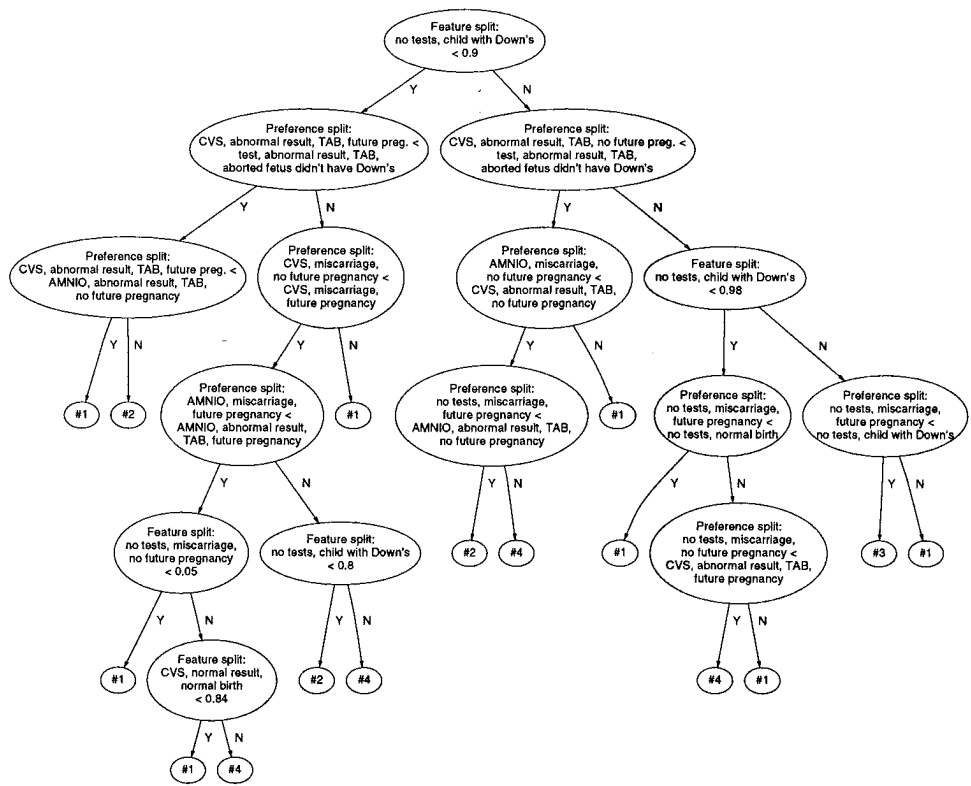

Figure 4: Decision tree for the TEEN history. The numbers at the leaves represent cluster labels and corresponding strategies (see Figure 3).

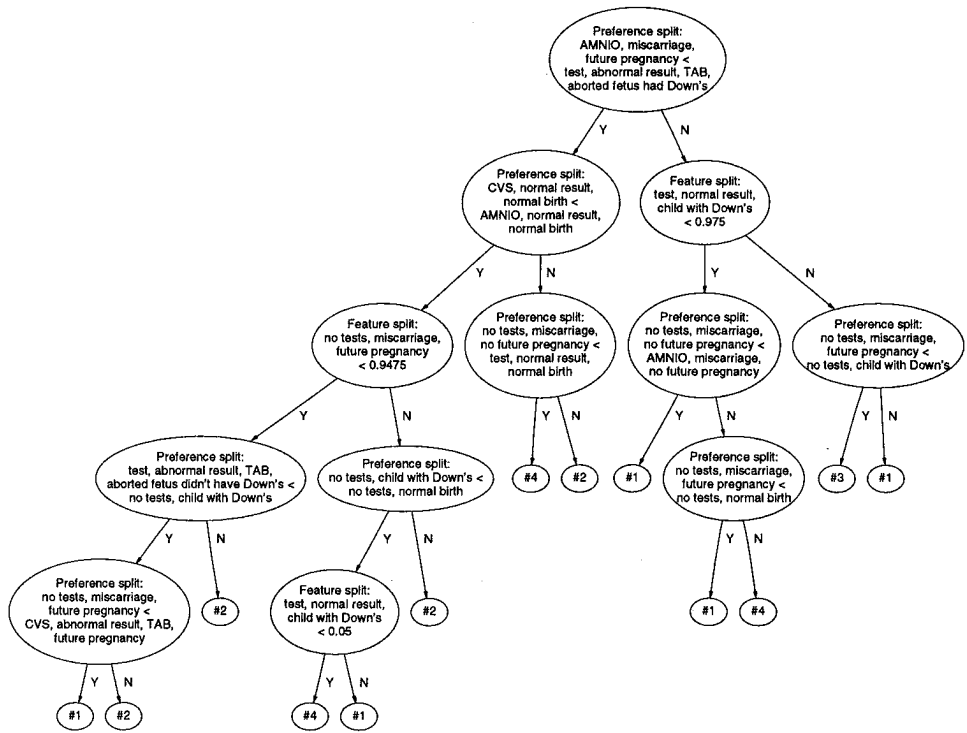

Figure 5: Decision tree for the 45YO history. The numbers at the leaves represent cluster labels and corresponding strategies (see Figure 3).

### 5.1 PROPERTIES OF THE DOMAIN

We began by computing for every utility function $u_j$ in our database the optimal strategy for the current history $h_k$: $s^*_{u_j|h_k}$. We also computed the expected utility of each of these strategies: $EU_{u_j}(s^*_{u_j|h_k})$. Now, for each utility function we compared the expected utility of its best strategy to the expected utility of the best strategy for each of the other utility functions. This allowed us to compute the UL for all pairs of utility functions.

We next clustered the data using our hierarchical agglomerative clustering algorithm. We found that for every age group we considered, we could identify four clusters with intra-cluster distances of 0. Thus, out of 18 strategies allowed in our model[3] only four strategies were found to be optimal for some utility function in the database. These four strategies are shown in Figure 3. Interestingly, the strategies didn't differ across the age groups. There were, however, significant differences in the sizes of clusters corresponding to these strategies. For some utility functions, the same strategy was optimal regardless of history. For others, it was different in every age group. These results correspond to the intuitions of many practitioners in the field of prenatal diagnosis. Since the probability of having a child with Down's syndrome increases dramatically with the age of the mother, for some women the optimal strategy should change with age. On the other hand, if the woman's attitude towards the possibility of having a disabled child is extreme (very negative or very positive), one strategy might be optimal for her throughout her life.

It is hard to characterize precisely the group of women for which a given strategy is optimal. The utility functions are sometimes very diverse even within the same cluster. Intuitively, we can think of Strategy 1 as appropriate for women with relatively low risk aversion towards the possibility of giving birth to a child with Down's syndrome. Strategy 4 should be used for women very risk averse in this respect. Strategy 2 is best for women who are equally afraid of having a child with Down's syndrome and aborting a healthy fetus. Strategy 3 was found to be appropriate for a woman who was also very afraid of aborting a healthy fetus and in addition, placed a very high value on knowing early in the pregnancy whether the disease was present. These characterizations, however, are very crude approximations.

After the data were clustered, we ran our decision tree algorithm. The trees generated for different histories were very different. Figures 4 and 5 show examples of decision trees for two extreme age groups: TEEN and 45YO. As we expected, the most important split for the TEEN tree was the feature split on deciding not to undergo any diagnostic procedure and having a Down's syndrome child. The probability of having a child with this disease is so low for this age group that only extreme values for this feature can cause the strategy to differ from the one the doctors usually advise their teen patients to pursue: "do nothing," i.e., strategy 1. Note that there are three feature splits for this feature in the tree, with different split points. On the other hand, the decision tree for 45YO demonstrates sensitivity to women's attitudes towards abortion and miscarriage. This phenomenon is again understandable, since the risk of Down's is the highest for this age group and all of the diagnostic tests carry a significant risk of inducing miscarriages.

Note that we need at most 6 (in the TEEN tree) or 5 (in the 45YO tree) assessments to completely identify the user's utility function. The numbers in the leaves of the decision tree represent different clusters and corresponding optimal strategies (see Figure 3). Recall that a full utility assessment for our model would require the user to assign values to 22 outcomes.

### 5.2 PERFORMANCE RESULTS

We were interested to see how well our algorithm would perform at classifying a new user. We measured the error of our algorithm by creating clusters and decision trees for a subset of the database, the training set, and then evaluated the performance of the decision tree on a small test set. We measure the error as the UL from using the utility function assigned by the decision tree as compared to the true utility function.

Figure 6 shows a sample learning curve for this domain. It shows the error as a function of the dataset size. We see that surprisingly, even for this small dataset, the results are quite promising. We are able to notice a steady decrease in error as the training set size increases. The results are particularly promising for the TEEN, 25YO and 35YO categories. The errors for AVE and 45YO are higher. A plausible explanation for the higher error for the 45YO users is that the choice of strategy is more sensitive for this age group to small changes in the utility function. This sensitivity is caused by the fact that very unlikely outcomes such as Down's or miscarriage are more probable for 45 year old women than for other age groups.

Figure 7 shows the error as a function of the number of clusters. We see that initially, as cluster size increases, we have an improvement in error. Then, as cluster size increases further, the error increases. The interesting phenomenon that we notice, is that despite the fact that there are only four strategies, it is actually better to create more than four clusters. This is because our decision trees are built over the space of the utility functions, while our clustering algorithm is geared toward identifying utility functions with similar optimal strategies. This curve also illustrates the classic phenomenon of overfitting—with too weak a bias (by allowing a large number of clusters), we overfit the data and observe poor performance on the test set. Using our algorithm, we can find the appropriate number of clusters.

There are several directions in which we would like expand our experiments. The most important of these is to test on a larger database for a richer model. Data for a larger study is currently being collected, and we plan to apply our algorithm to this data when it becomes available.

---
[3]We restricted the space of possible strategies to eliminate nonsensical ones, e.g., involving prenatal tests after termination.

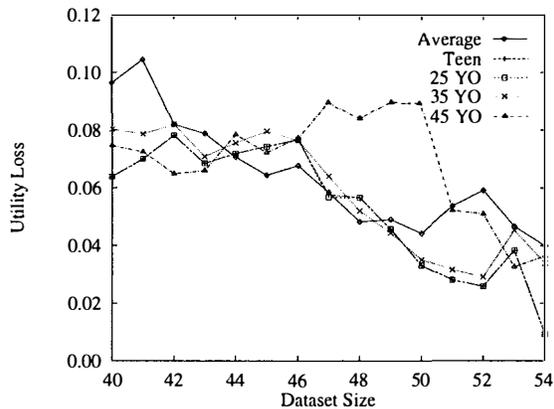

Figure 6: Learning curves (average of 10,000 runs).

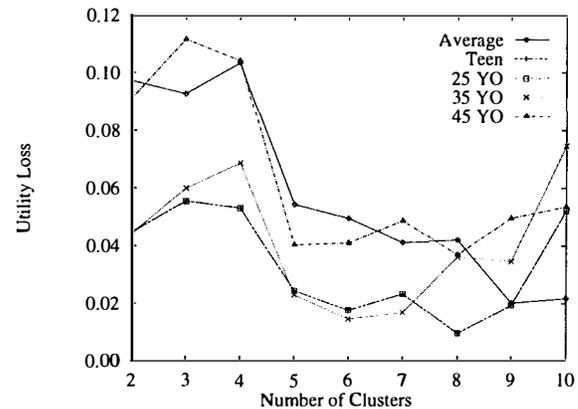

Figure 7: Leave-one-out cross-validation for number of clusters.

## 6 RELATED WORK

With recent advances in the power and scope of decision-theoretic systems, utility elicitation has become a lively and expanding area of research. A few projects are particularly relevant to this work. Hornberger (1995) applied CART classification to utilities for the purpose of simplifying utility assessment. His work assumed linear additive independence between utility attributes, and used simulated rather than observed utilities as a basis for classification. Horvitz and Klein (1993) used utility as a basis for categorization in the context of decision analysis. They abstract policy and outcome spaces using clustering approaches that are similar to ours. Farr and Shachter (1992) designed a system to assess utilities parsimoniously using simulated decision scenarios rather than complete standard-gamble utility assessment. Their approach was geared toward eliciting utilities from a single expert user rather than many individual end-users. Finally, Jimison (1992) addressed the broader task of tailoring general decision models to individual users by explicitly representing uncertainties about key utility and probability parameters. Her work was geared toward explaining clinical decision models and refining them in a principled way using expected value of information.

## 7 CONCLUSIONS AND FUTURE WORK

We have presented a new approach to utility function elicitation based on machine learning techniques. This method provides an effective alternative to the current approaches to utility elicitation. We neither have to make assumptions about the decomposability of the utility function nor are we limited to eliciting only preferences; thus our approach should suit a wide variety of applications. We have applied this framework to a prenatal testing decision model, and found the results quite promising. We found that we could in fact identify a small number of prototype utility functions. We also found that we could classify a new user's utility function with only a small number of easy-to-answer questions, with the introduction of only minimal error due to using the prototype utility function rather than the user's true utility function.

The output of our approach, a decision tree, has the advantage of being easily interpretable by practitioners. More importantly, it makes the elicitation of user preferences much easier. The types of questions the users must answer are simpler than standard gamble questions. We hope that this will enable individualized utility elicitation for users in real clinical settings.

We plan to continue our research in several directions:

- We would like to bound the UL for a particular patient given the cluster we have identified. Every leaf in our decision tree corresponds to a convex region in the utility space. By looking at the optimal policies at the vertices of this region, we can come up with a bound on the UL over the entire region. In cases where this bound is not tight enough, we may ask additional questions until we achieve satisfactory bounds.

- We would like to explore the possibility of including history information in the clustering procedure. Currently we create separate clusterings for each history, using the entire database under the assumption that the utility function is independent of the history. This assumption allows us to avoid data segmentation in our small database. We would like to explore the consequences of relaxing this assumption.

- Currently, we do not consider clustering based on the similarity of the actions in the policies, we only cluster on the basis of UL. We also do not consider abstractions of the outcome space. It would be quite interesting to consider flexible methods of clustering along all three of these dimensions.

- While many utility functions are not decomposable in general, we may find a decomposition applicable to particular clusters. We could also use conditional independence between the attributes to speed up the classification process.

- In many cases, there are some obvious constraints that all utility functions should obey. For example, in the PANDA model, we can assume that a woman seeking advice on the choice of diagnostic tests would consider having a healthy baby more desirable than having a baby with a severe disability. We would like to explore the possibility of using such constraints—background knowledge—to increase the effectiveness of our procedure.

**Acknowledgments**

We are very grateful to Miriam Kuppermann of UCSF/Mount Zion Medical Center for sharing her database of utility functions. We would like to thank Mehran Sahami for useful suggestions and use of his document clustering code that we adapted to our framework. We also used the inference engine made available to us by Cecil Huang. We discussed the work presented in this paper with many people. We'd like to acknowledge helpful comments from Xavier Boyen, Denise Draper, Jerome Friedman, Eric Horvitz, Daphne Koller, Mark Peot and Yoav Shoham. Samuel Posner helped us with the transmission and format of the database. Urszula Chajewska was supported by the ARPI grant F30602-95-C-0251. Lise Getoor was supported by a National Physical Sciences Consortium fellowship. This work was also supported through the generosity of the Powell Foundation, by ONR grant N00014-96-1-0718, and by DARPA contract DACA76-93-C-0025, under subcontract to Information Extraction and Transport, Inc. Joseph Norman was supported by the Medical Scientist Training Program, and Yuval Shahar by National Library of Medicine grant LM06245 and National Science Foundation grant IRI-9528444.